\documentclass[sigconf,10pt]{acmart}

\usepackage{amssymb}

\usepackage{cite}
\usepackage{amsmath,amssymb,amsthm}
\usepackage{multicol}
\usepackage{multirow}
\usepackage{bm}
\usepackage{setspace} 
\usepackage{algorithm}
\usepackage{algpseudocode}
\usepackage{array}
\usepackage{subfigure}
\AtBeginDocument{%
  }

  \acmConference[XXX ’XX]{XXX,XXX
  2024}{XXX,XXX}

\begin{document}

\title{DIVESPOT: Depth Integrated Volume Estimation of Pile of Things Based on Point Cloud}

\author{Yiran Ling}
\email{2021110742@stu.hit.edu.cn}

\affiliation{
  \institution{Harbin Institute of Technology}
  \city{Harbin}
  \state{Heilongjiang}
  \country{China}
}

\author{Rongqiang Zhao}
\authornote{Corresponding author.}
\email{zhaorq@hit.edu.cn}
\affiliation{
  \institution{Harbin Institute of Technology}
  \city{Harbin}
  \state{Heilongjiang}
  \country{China}
}

\author{Yixuan Shen}
\email{2021113638@stu.hit.edu.cn}
\affiliation{
  \institution{Harbin Institute of Technology}
  \city{Harbin}
  \state{Heilongjiang}
  \country{China}
}

\author{Dongbo Li}
\email{ldb@hit.edu.cn}
\affiliation{
  \institution{Harbin Institute of Technology}
  \city{Harbin}
  \state{Heilongjiang}
  \country{China}
}

\author{Jing Jin}
\email{jinjinghit@hit.edu.cn}
\affiliation{
  \institution{Harbin Institute of Technology}
  \city{Harbin}
  \state{Heilongjiang}
  \country{China}
}

\author{Jie Liu}
\email{jieliu@hit.edu.cn}
\affiliation{
  \institution{Harbin Institute of Technology}
  \city{Harbin}
  \state{Heilongjiang}
  \country{China}
}


\begin{abstract}
Non-contact volume estimation of pile-type objects has considerable potential in industrial scenarios, including grain, coal, mining, and stone materials. However, using existing method for these scenarios is challenged by unstable measurement poses, significant light interference, the difficulty of training data collection, and the computational burden brought by large piles. To address the above issues, we propose the Depth Integrated Volume Estimation of Pile Of Things (DIVESPOT) based on point cloud technology in this study. For the challenges of unstable measurement poses, the point cloud pose correction and filtering algorithm is designed based on the Random Sample Consensus (RANSAC) and the Hierarchical Density-Based Spatial Clustering of Applications with Noise (HDBSCAN). To cope with light interference and to avoid the relying on training data, the height-distribution-based ground feature extraction algorithm is proposed to achieve RGB-independent. To reduce the computational burden, the storage space optimizing strategy is developed, such that accurate estimation can be acquired by using compressed voxels. Experimental results demonstrate that the DIVESPOT method enables non-data-driven, RGB-independent segmentation of pile point clouds, maintaining a volume calculation relative error within 2\%. Even with 90\% compression of the voxel mesh, the average error of the results can be under 3\%.

\end{abstract}





\maketitle

\section{INTRODUCTION}
Point cloud volume calculation technology reconstructs three-dime-nsional objects by acquiring point clouds through camera perception, depicting discrete coordinate information of the object's surface, and calculating the object's volume based on this information. This technology features non-contact operation, low cost, a wide measurement range, and ease of data processing, making it widely applicable in cross-scale irregular object volume calculations, ranging from fruits, flowers, and toys to statues, stone piles, and mountains\citep{Ding_2023}\citep{wang2023calculation}\citep{leite2020estimating}\citep{griwodz2021alicevision}\citep{ZHAO2021106131}.

In industrial volume measurement scenarios, pile-type objects have consistently presented measurement challenges\citep{WOS:000425915900015}\citep{rs11060623}. While using weighbridges or conveyor belts for small-scale, batch measurements can partially meet the measurement demands, the equipment, labor, and time costs often render these methods impractical for efficient automated deployment. Volume measurement based on point cloud 3D reconstruction\citep{rs15205006} can largely fulfill the needs of such scenarios by estimating through sparse 3D reconstruction information. However, current research on algorithms in this area is still inadequate, making it difficult to achieve full-process automated deployment. Additionally, the estimation accuracy often falls short of practical application requirements.

Current principles of point cloud volume estimation can be divided into three categories: slice-based methods\citep{zhou2021research}\citep{isprs-archives-XL-5-101-2014}\citep{10044532}, voxel-based methods, and convex hull-based methods\citep{stefanidou2020lidar}\citep{doi:10.1080/01431161.2018.1541111}. The slice-based method involves slicing the object's point cloud into layers, counting the points in each slice, and integrating the volume of these local slices to calculate the total volume. The voxel-based method\citep{moreno2020ground} extends the concept of calculating two-dimensional areas to three-dimensional volumes by dividing the three-dimensional space into regular 3D voxel grids and counting the globally consistent voxels to compute the volume. The convex hull-based method constructs a convex hull structure around the point cloud, calculating the volume inside the convex hull to obtain the point cloud's volume.The 3D convex hull, constructed using the $\alpha$-shape method, calculates an object's volume by summing the positive and negative volumes of triangular prisms projected onto a plane, while the 2D convex hull optimizes the shape of sliced layers and calculates their area.

However, the current process of forming a complete measurement scheme from the principles of point cloud volume estimation\citep{rs15205006} faces many challenges, restricting the accuracy of practical volume calculations and making it difficult to meet the requirements for widespread industrial and everyday use\citep{hongyan2021volume}\citep{bin2019slicing}.In the implementation of volume calculation principles, the slice-based method's accuracy is influenced by the number of slices, where more slices reduce computational efficiency and increase shape information loss, complicating optimization, while higher slice layers preserve shape information but decrease height direction accuracy. The convex hull method, using techniques like $\alpha$-shape for densification, often requires manual hyperparameter adjustments, resulting in excessive smoothing or incomplete local densification on irregular objects, thus compromising robustness.

\begin{figure}[h]
    \centering
    \includegraphics[width=\linewidth]{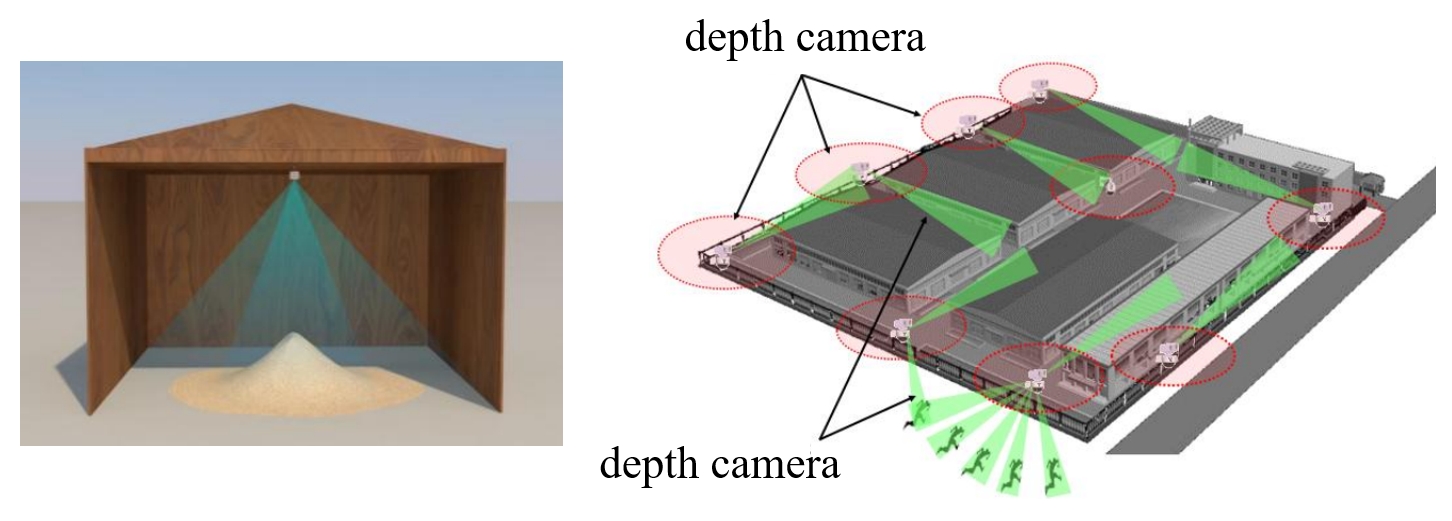} 
    \caption{Demonstration of DIVESPOT configured in grain silos.}
    \label{fig:grainbin}
\end{figure}

Moreover, regarding the quality of the point cloud itself, the initial acquisition and construction of the point cloud are prone to outlier noise points due to the randomness of algorithms in steps such as registration\citep{kang2020review}\citep{CHEN2024103683}\citep{10422189}\citep{10160307}. As demostrated in Fig. \ref{fig:grainbin},The surrounding environment of the object also contains extensive interfering noise points, which can affect subsequent computational algorithms to varying degrees. Although current semantic-based point cloud segmentation technology can reliably extract the target object for denoising and object extraction\citep{wang2020estimation}\citep{ruchay2021fast}, it often relies on extensively manually annotated datasets and indoor scenes. The arbitrary shapes and geometric constraints of irregular industrial pile objects, combined with sparse dataset features, significantly limit the effectiveness of point cloud segmentation.

Additionally, current point cloud volume calculation schemes lack calibration and correction for the original point cloud, making it difficult to determine the overall pose of the point cloud\citep{https://doi.org/10.1155/2024/6662678}.\citep{li2016dual} This challenge hinders the implementation of automated translation and segmentation algorithms for the point cloud, increasing the difficulty of denoising and object extraction and introducing more interference from other objects into the computational algorithms\citep{ZHAO2021106131}. As a result, point cloud volume calculations are often confined to fixed camera setups and manual denoising in single scenarios, limiting their applicability to larger and more diverse environments.

In our study, corn grain piles serve as the experimental subject, with RGB-D point cloud data collected through a Time-of-Flight (ToF) depth camera to complete the 3D reconstruction. Given that the application scenarios for this technology are typically open or semi-open warehouse environments, which contain numerous interfering point cloud objects and require measurements at different times of the day, we introduce various interfering obstacles in the surroundings. Additionally, we ensure that no information related to RGB data is used in the volume measurement calculation algorithm. Under these background settings and constraints, we design and conduct research on the core algorithms of the measurement system.

This study addresses the limitations of existing research and makes the following contributions:

1. We designed DIVESPOT, which includes pose calibration, automated filtering, and a voxel network-based volume estimation algorithm. This system successfully addresses the challenges of pose uncertainty and large-volume noise point clouds in point cloud-based pile volume measurement.

2. We proposed a ground feature extraction algorithm based on point cloud height density distribution. Using this algorithm, we achieved non-data-driven, RGB-independent segmentation of pile point clouds. Under fully automated implementation, the volume measurement in outdoor, irregular pile scenarios achieved a relative error accuracy within 2\%.

3. We optimized the storage space efficiency of the algorithm, ensuring high estimation accuracy even with a 90\% voxel network compression rate, and validated the performance on typical embedded devices.

\section{MOTIVATION AND CHALLENGE}

\begin{figure*}[h]
    \centering
    \includegraphics[width=0.8\textwidth]{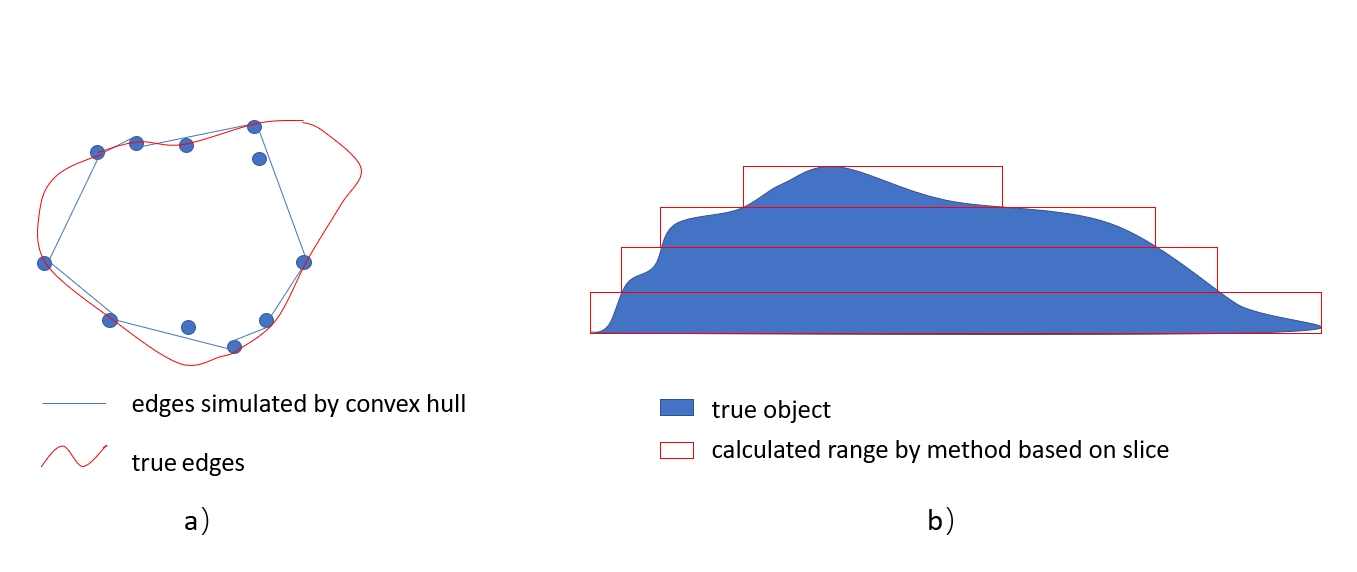} 
    \caption{Slice-based methods a) When the interval length is taken too small,it will result in small individual slice contour predictions .b) When the interval length is too large,it will result in an extra increment in the integral  making the volume large.}
    \label{fig:badslice}
\end{figure*}

The challenges in point cloud volume measurement scheme for pile objects arise from various aspects. The design of the entire measurement scheme should fully consider several factors: improvements in the principles of existing point cloud volume estimation algorithms, physical constraints from the application scenarios, and considerations for the practical deployment of the scheme, as well as the design of various additional algorithmic components necessary for full-process automated deployment.

Currently, the commonly used point cloud measurement principles are mainly based on slice and $ \alpha$-shape methods. The limitations of these methods in practical applications are their lack of robustness and reliance on hyperparameters, preventing adaptive adjustment across different point cloud densities and pile shapes to ensure the reliability of measurement results.

\begin{figure}[h]
    \centering
    \includegraphics[width=0.9\linewidth]{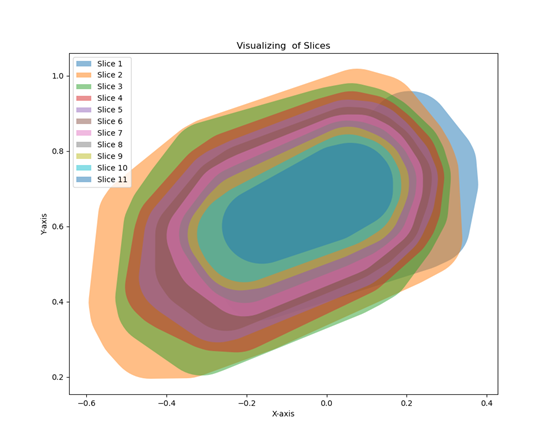} 
    \caption{Demonstration of Slice-based volume calculation, obtained by looking down from above the pile.}
    \label{fig:slicedemo}
\end{figure}

\begin{figure}[h]
    \centering
    \includegraphics[width=0.8\linewidth]{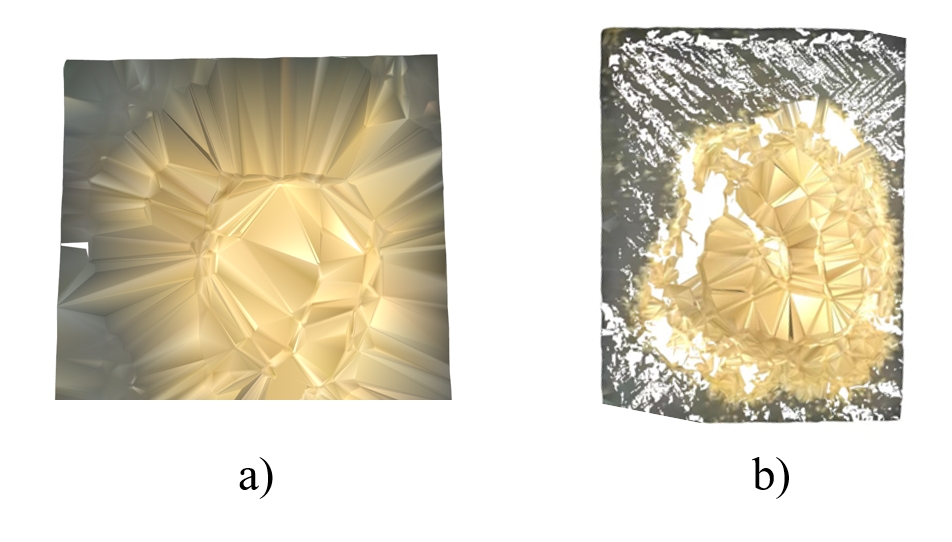} 
    \caption{Convex packet-based volume calculation methods rely on hyperparameters and tend to lead to a)loss of information or b)broken reconstruction.}
    \label{fig:convex}
\end{figure}

The slice-based method, as demonstrated in Fig. \ref{fig:slicedemo},faces significant challenges when dealing with low-density point clouds. The sparse point cloud within a single slice layer results in a significant reduction in the accuracy of area calculations, making it difficult to accurately fit the true point cloud contour and thereby reducing the final calculation accuracy. Consequently, it is difficult to set very thin slice layers in these algorithms, inherently limiting their maximum accuracy. Conversely, reducing the number of slices leads to thicker single slices, which, while containing more information for better fitting, introduces noticeable volume calculation errors due to the larger step size during integration. This is particularly problematic for piles with small slope angles, where maintaining calculation accuracy is challenging. As illustrated in Fig. \ref{fig:badslice},such algorithms thus require highly complex slice contour fitting based on pile semantics and have high demands on the initial point cloud density. The convex hull method, whether two-dimensional or three-dimensional, heavily depends on the $\alpha$ value. A large $\alpha$ value results in a coarse fit, as shown in Fig. \ref{fig:convex}rendering the shape inaccurate for high-precision volume estimation. On the other hand, achieving a fine fit for the target shape introduces significant algorithmic complexity due to the identification and transformation required for convex and concave hull applications, leading to poor robustness and unsuitability for edge deployment.

\begin{figure*}[h]
    \centering
    \includegraphics[width=0.8\linewidth]{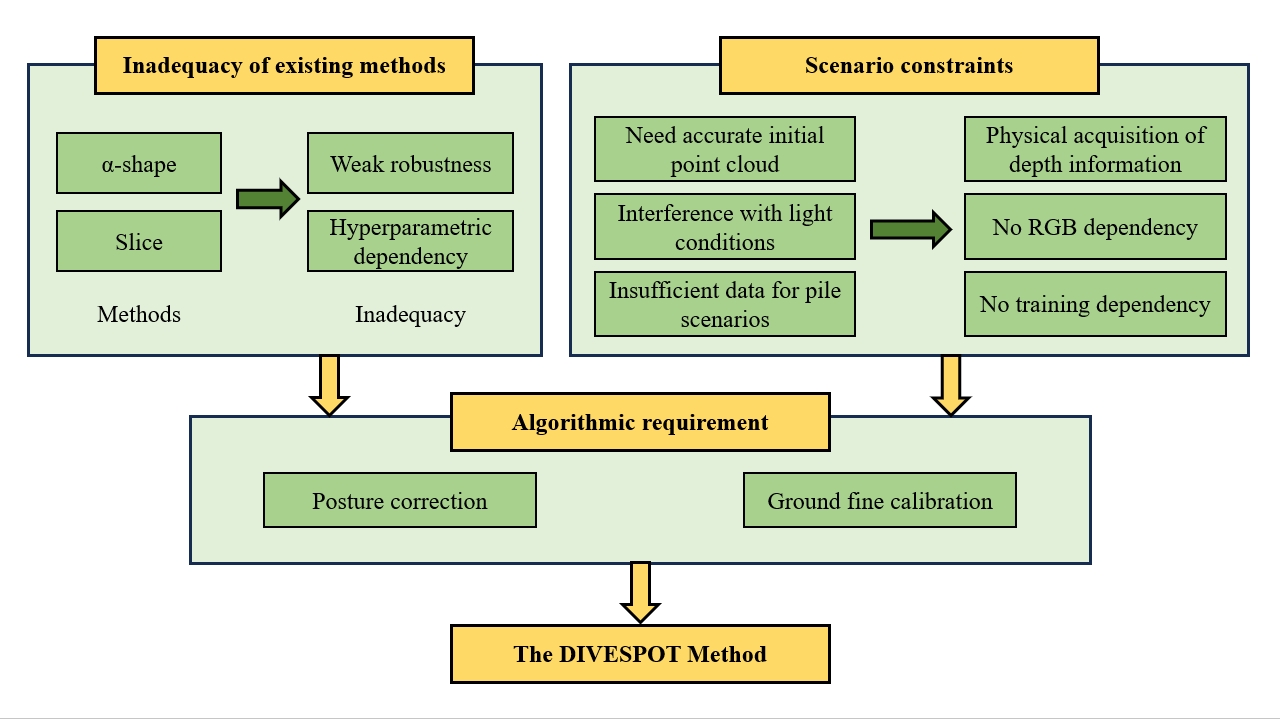} 
    \caption{Motivation and challenge of DIVESPOT}
    \label{fig:motivation}
\end{figure*}

Therefore,as summarised in Fig.\ref{fig:motivation} , more straightforward method is desired, which directly calculates volume based on the acquired point cloud or voxel network. While this requires more algorithmic post-processing and may demand more memory space for performance, it ensures the full utilization of the original point cloud data.

From the perspective of practical application requirements, three key constraints for the design of the point cloud measurement system can be identified. To achieve a measurement error of 5\% or even below 2\%, it is essential to obtain the most accurate original point cloud possible. Considering that piles are typically placed in semi-open environments, a rational choice is to use depth cameras to directly capture 3D point cloud data. Since measurements may occur both during the day and at night, the impact of insufficient lighting must be considered. Therefore, active ToF cameras or LiDAR should be used, which further implies that the algorithm should not depend on RGB information, as the point cloud might lack reliable RGB attributes.

In addition to these two factors, the physical shape characteristics of the pile itself also constrain the system algorithm design. Piles are composed of numerous small particles, with an irregular overall shape and no fixed paradigm, only a local angle determined by the shape of the particles and gravity. This makes it challenging to construct a dataset that fully represents their characteristics while meeting economic requirements. Hence, the feasibility of training a neural network to perform semantic segmentation tasks based solely on shape features is limited.

For the automated deployment of the overall scheme, additional post-processing of the point cloud must be considered. Firstly, to calculate the volume from the original point cloud, regardless of the calculation principle used, the pose of the point cloud must be accurately determined. Therefore, pose adjustment of the point cloud is required initially. 

In conjunction with the voxel network-based volume measurement algorithm, ground calibration is necessary. A lightweight algorithm must be designed to accurately extract ground features. In actual scenarios, many noise points exist, and the point cloud acquisition algorithm introduces noise points due to registration and other reasons. Particularly when there are people, large machinery, and structures in the surroundings, substantial noise and interfering point clouds are often present. 

Therefore, a point cloud filtering algorithm needs to be designed. This algorithm should perform initial trimming of the point cloud calculation scene and further filter out the remaining noise points.

\section{METHOD}

\begin{figure*}[t]
    \centering
    \includegraphics[width=0.8\linewidth]{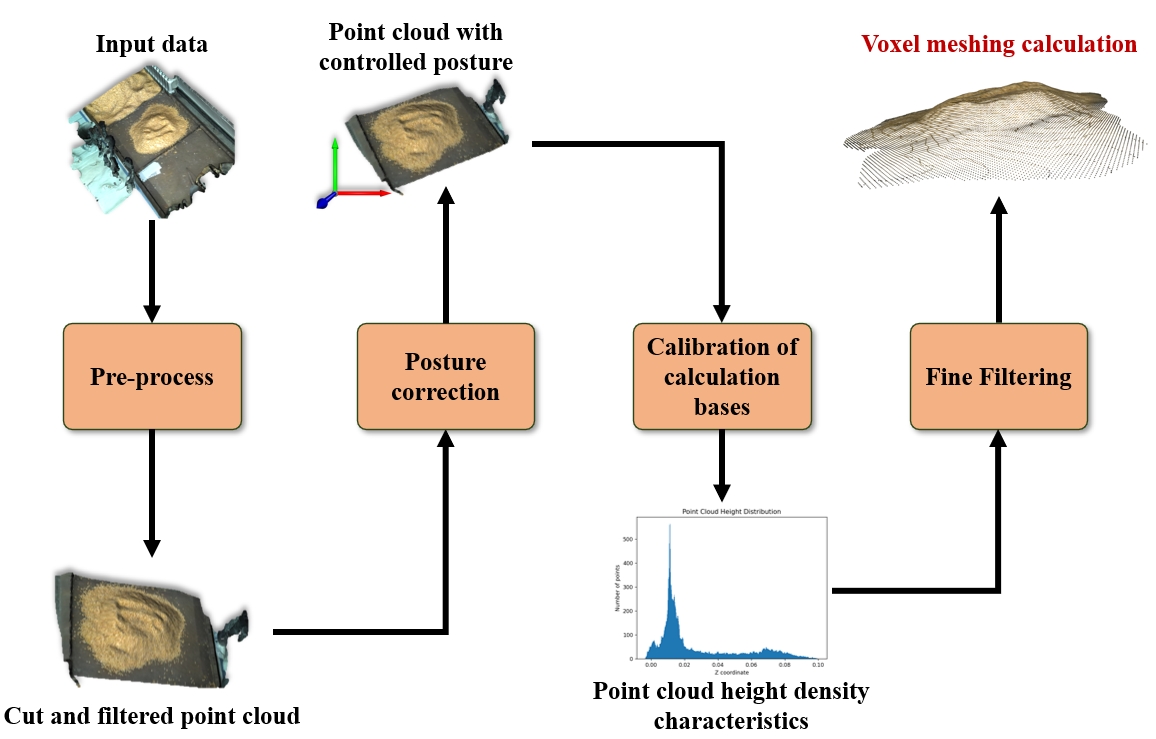}
    \caption{Overview of DIVESPOT}
    \label{fig:overview}
\end{figure*}

\textbf{Overview}: The workflow of DIVESPOT is illustrated in Fig.\ref{fig:overview}. The process comprises five fundamental parts. After obtaining the original scene reconstruction point cloud, Pre-process removes redundant environmental objects and noise from the surroundings. Next, Posture Correction aligns the point cloud's pose with the horizontal plane of the world coordinate system. Calibration of the calculation basis involves analyzing the height density histogram to determine the position of the reference plane. Fine Filtering makes final adjustments and further filters the point cloud. Finally, volume calculation is performed using the voxel meshing calculation method.

\textbf{Pre-process}: The acquired scene point cloud undergoes three stages of pre-processing: 1) initial point cloud trimming, 2) noise filtering, and 3) voxelization for uniform sampling. The initial point cloud contains numerous peripheral interfering objects and noise points generated during point cloud creation. The first step in pre-processing involves trimming the scene. Given the uniformity of the camera measurement poses, the point cloud poses are also relatively uniform. Based on prior estimation or recognition of the measurement position, pass-through filtering is applied to the coordinate region of the point cloud. Subsequently, noise filtering is performed on the preliminary trimmed point cloud using radius filtering combined with the HDBSCAN algorithm\citep{campello2013density}, as detailed in Algorithm 1. we use:
\begin{equation}
P_{(m \times 3)} = \begin{pmatrix}
x_1 & y_1 & z_1 \\
\vdots & \vdots & \vdots \\
x_m & y_m & z_m
\end{pmatrix}
\end{equation}
to represent the 3D coordinate matrix of the point cloud, where the row vectors represent the three coordinates and the column vectors represent the different points. And $\mathcal{H}$ represents the HDBSCAN-based algorithm,Clustering algorithm that adaptively find the most classes after clustering. After initial pre-processing, voxelization is applied to uniformly sample the point cloud. Assuming uniform point cloud sampling, the projection of each point cloud on the ground is also uniform. The area of each point cloud's corresponding projection element can be obtained by dividing the known scene area by the number of point clouds, yielding the area of each point cloud's corresponding projection element.

\begin{algorithm}
\caption{Highly robust point cloud filtering}
\begin{algorithmic}[1]
\State \textbf{Input:} Point cloud matrix that needs filtering $P_{m \times 3}$; Filter radius $r_0$; $n_{\text{min}}$
\State $P_{\text{filtered}}  \leftarrow \emptyset$
\For{$i = 1$ $\mathbf{to}$ $N$}
    \State $n \leftarrow 0$
    \State $\text{point}_i \leftarrow P[i, 3]$
    \For{$j = 1$ $\mathbf{to}$ $N$}
        \If{$i \neq j$}
            \State $d \leftarrow \|\text{point}_i - P[j, 3]\|$
            \If{$d \leq r_0$}
                \State $n \leftarrow n + 1$
            \EndIf
        \EndIf
    \EndFor
    \If{$n \geq n_{\text{min}}$}
        \State $P_{\text{filtered}}.\text{append}(\text{point}_i)$
    \EndIf
\EndFor
\State $P_{\text{filtered}} \leftarrow \mathcal{H}(P_{\text{filtered}})$
\State \textbf{Output:} $P_{\text{filtered}}$
\end{algorithmic}
\end{algorithm}

\textbf{Posture correction}: After pre-processing, the point cloud's actual pose and position still contain uncontrollable factors due to direct acquisition. To accurately calculate the volume, pose correction is necessary to align the point cloud's coordinate system with the world coordinate system. We use a RANSAC-based\citep{fischler1981random} recognition algorithm to identify the most densely populated ground points in terms of point cloud semantics. Then, we use the Rodrigues rotation formula to rotate the point cloud based on the ground's normal vector to achieve a horizontal pose,and preliminarily translate it to the XOY plane of the world coordinate system:
\begin{equation}
\mathbf{v}_{\text{rot}} = \mathbf{v} \cos \theta + (\hat{e_z} \times \mathbf{v}) \sin \theta + \hat{e_z} (\hat{e_z} \cdot \mathbf{v}) (1 - \cos \theta)
\end{equation}

The detailed algorithm workflow is provided in Algorithm 2. $\mathcal{R}$  represents using the RANSAC algorithm to fit the set of points that most resemble the plane. $\mathcal{N}$ represents to normalize the value. $\boldsymbol{r}$represents calculating the rotation matrix between two vectors using the Rodrigues rotation formula.

\begin{algorithm}
\caption{Point cloud posture correction}
\begin{algorithmic}[1]
\State \textbf{Input:} Point cloud matrix after straight-through filtering $P_{m \times 3}$
\State $P_{\text{plane}} \leftarrow \mathcal{R}(P_{\text{origin}})$
\State Compute $A, B, C, D$ for platform $\alpha:$ $ A x + B y + C z + D = 0$ $\boldsymbol{s.t.}$ $\min(d_{P_\alpha})$
\State Unit normal vector: $\boldsymbol{v} \leftarrow \mathcal{N}(A, B, C)$
\State Rotation matrix: $\boldsymbol{T} \leftarrow \boldsymbol{r}(\hat{e_z}, \boldsymbol{v})$
\State $P_{\text{rotated}} \leftarrow P_{\text{origin}}\boldsymbol{T}$
\State \textbf{Output:} Point cloud set after rotation $P_{\text{rotated}}$
\end{algorithmic}
\end{algorithm}
\begin{figure}[h]
    \centering
    \includegraphics[width=\linewidth]{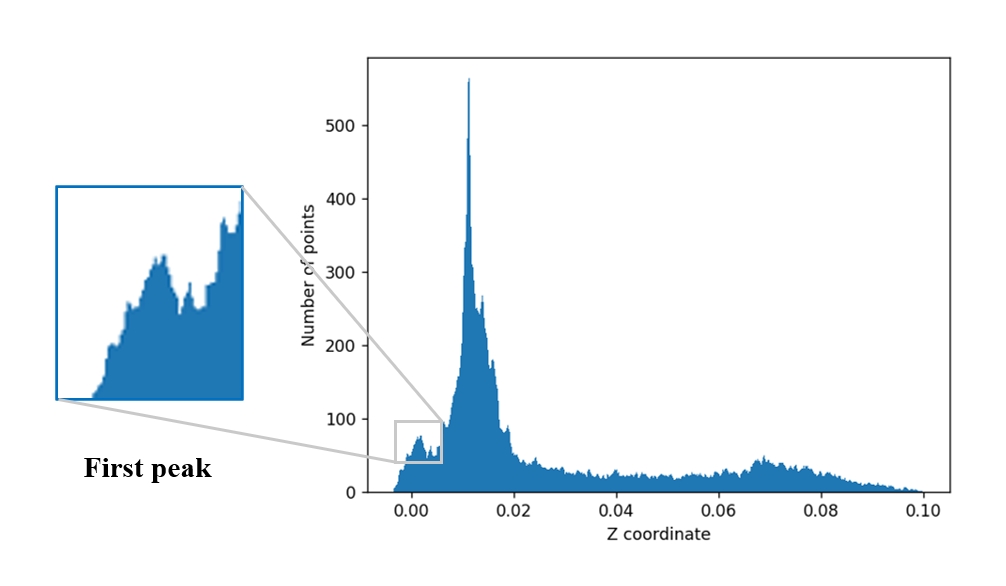}
    \caption{After filtering the height density figure, select the first peak from left to right as the ground position.}
    \label{fig:firstpeak}
\end{figure}
\textbf{Calibration of calculation basis}: After the preliminary translation, further refinement is necessary for precise volume measurement by finely calibrating the measurement reference plane (usually the ground). We propose a ground determination scheme based on height density distribution maps, with the specific code implementation workflow shown in Algorithm 3.,where $\mathcal{SF}$ represents Straight Filtering,the simple filtering method to filter all point clouds above a given height value. As shown in Fig.\ref{fig:firstpeak}, the horizontal axis of the height density image represents the point cloud height distribution intervals, and the vertical axis represents the number of point clouds within each interval. Different intervals yield characteristic information of the point cloud in the height direction. Since the semantic information of the measurement reference ground essentially comes from its continuous horizontal plane, this non-recognition algorithm method can directly and uniformly measure the height of the entire point cloud in the height direction. Considering the random errors caused during point cloud registration, the ground height will exhibit a Gaussian-like distribution. We might as well take the horizontal coordinate at the first corresponding vertical coordinate peak from left to right as the position of the ground height to ensure consistency of the ground position across different images. In practical calculations, an empirical interval is taken from low height values, and the height corresponding to the maximum number of points within this interval is determined.

\begin{algorithm}
\caption{Ground extraction using Height density distribution}
\begin{algorithmic}[1]
\State \textbf{Input:} Point cloud matrix to be analyzed $P_{m \times 3}$; $n_{\text{interval}} \in \mathbb{N}$; $step$
\State $z  \leftarrow P[:, 3]^T$
\State Construct $n_{\text{interval}}$ intervals with $z_{\text{min}}, z_{\text{max}}$ as upper and lower bounds
\State $s_{\text{counts}}  \leftarrow [\text{z,intervals}]$
\State $n_{\text{start}}$ $ \leftarrow 0.5 \cdot step$
\State $n_{\text{end}}$ $ \leftarrow \text{length}(s_{\text{counts}}) - n_{\text{start}}$
\State $s_{\text{res}}  \leftarrow \emptyset$
\For{$i = n_{\text{start}}$ $\mathbf{to}$ $n_{\text{end}}$}
    \State $v$ $\leftarrow 0$
    \For{$j = i - n_{\text{start}}$ $\mathbf{to}$ $i + n_{\text{start}}$}
        \State $v$ $ \leftarrow v + s_{\text{counts}}[j]$
    \EndFor
    \State  $ v \leftarrow \frac{v}{step} $
    \State $s_{\text{res}}.\text{append}(v)$
\EndFor
\State $s_{\text{counts}} \leftarrow s_{\text{res}}$
\State $\mathbf{Find}$ the height represented by the maximum value in the lower empirical interval in counts $\mathbf{as}$ $ground$
\State $P_{\text{processed}}  \leftarrow \mathcal{SF}(P_{\text{origin}}, ground)$
\State \textbf{Output:} Point cloud set after analysis and filtering $P_{\text{processed}}$
\end{algorithmic}
\end{algorithm}

\textbf{Fine Filtering}: After calibrating the measurement reference plane, the point cloud needs to be further refined and translated according to the ground position. Additionally, parts below this calculation reference plane must be further filtered. Using pass-through filtering, the portions below the plane are filtered out. However, some ground point cloud residues may still remain, which can be further removed using radius filtering combined with the HDBSCAN algorithm.

\textbf{Voxel Meshing Calculation}: For volume calculation, we perform a grid division based on the ground. The area of each ground element has already been determined. By integrating the product of the element area and the height of the corrected point cloud, we can calculate the total volume of the point cloud. In practice, extracting the ground peak position in the height density image might encounter situations where the peak is not very distinct. Therefore, pass-through filtering based on a slightly higher ground position is often employed to ensure that near-ground noise points are filtered out as much as possible. However, this approach might also trim the bottom of the measured object, resulting in a smaller volume after the final integration step. To address this, we apply a volume compensation factor based on empirical data derived from the stacking characteristics of industrial measurement objects to adjust the volume calculation.

\section{EVALUATION}

\subsection{Volume Estimation Effectiveness Evaluation}
A volume estiomation evaluation method based on variable volume and shape of corn grain piles is constructed. The testing environment is an open indoor location, with static interfering objects such as support rods, walls, and miscellaneous piles around the corn heap, as well as dynamic obstacles created by moving personnel. This setup allows the effectiveness of the point cloud filtering algorithm to be tested based on the filtering of relevant obstacles. The shape of the corn heap is irregular and unpredictable, and due to the hardness of the grains and the difficulty in altering gaps between them, it is statistically considered that changing the heap configuration does not result in volume changes greater than the estimation accuracy. Based on these characteristics, varying the overall volume and shape of the corn heap as different groups of test objects ensures both the randomness of the data and the simplicity of the operations. Additionally, it ensures that the evaluation of our method's effectiveness possesses sufficient confidence and repeatability.

\begin{figure}[h]
    \centering
    \includegraphics[width=0.8\linewidth]{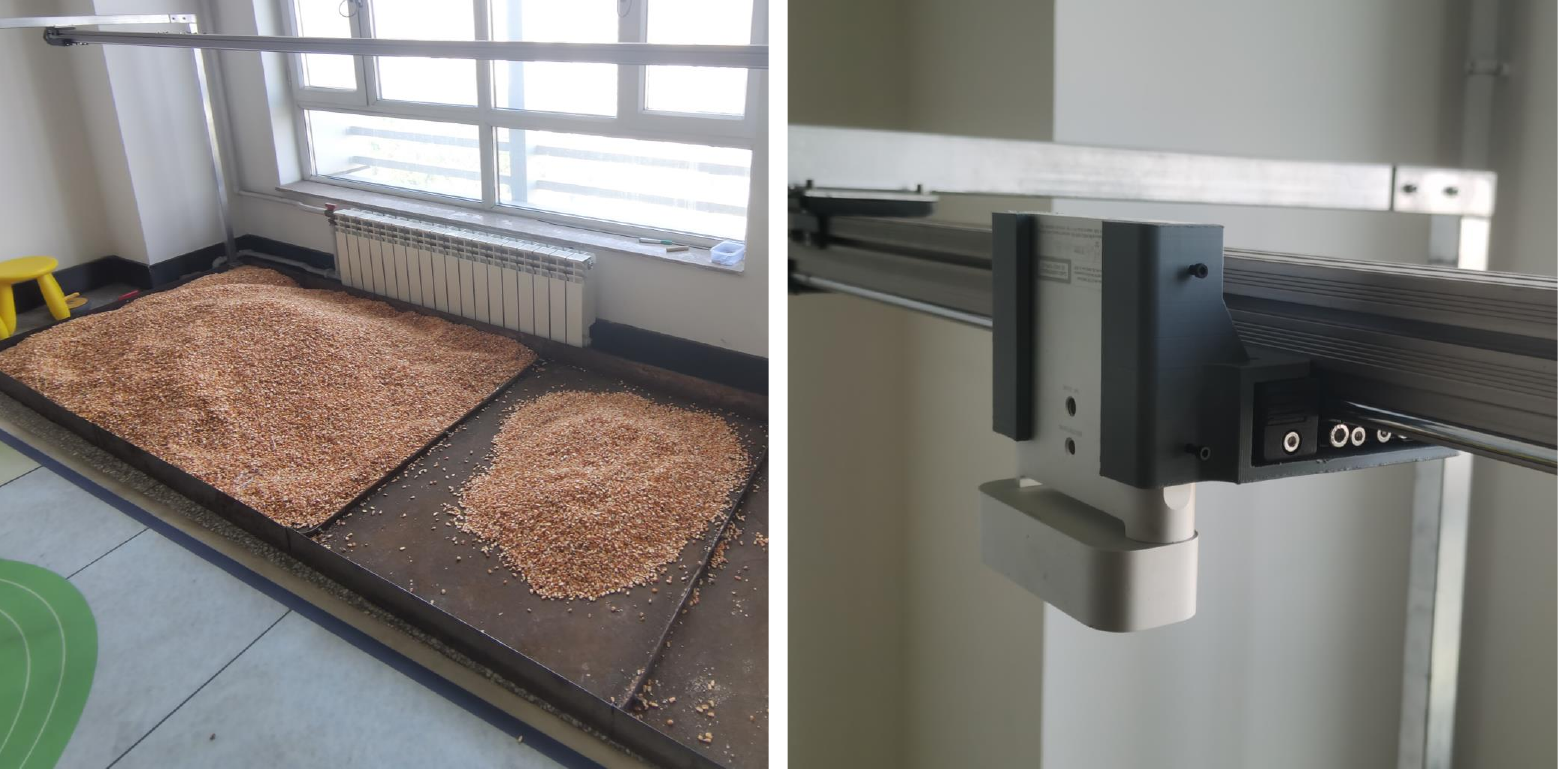}
    \caption{Experimental Scene Rigging and Depth Image Acquisition Camera.}
    \label{fig:RealPileAndCamera}
\end{figure}

\begin{table*}[ht]
    \centering
    \small
    \begin{tabular}{cccccc}
        \toprule
        \textbf{Area size/$m^2$} & \textbf{Real pile volume/$m^3$} & \textbf{Pile shape serial number} & \textbf{Estimated volume/$m^3$} & \textbf{50 rounds mean error} & \textbf{Mean error} \\
        \midrule
         & 0.014 & 1  & 0.01376 & 1.72\% &  \\
            & 0.014 & 2  & 0.01367 & 2.38\% &         \\
            & 0.014 & 3  & 0.01376 & 1.74\% &         \\
            & 0.028 & 4  & 0.02760 & 1.42\% &         \\
    1.3 & 0.028 & 5  & 0.02758 & 1.51\% &  1.97\%       \\
            & 0.028 & 6  & 0.02748 & 1.85\% &         \\
            & 0.035 & 7  & 0.03430 & 2.01\% &         \\
            & 0.035 & 8  & 0.03408 & 2.63\% &         \\
            & 0.035 & 9  & 0.03414 & 2.47\% &         \\
            \hline
         & 0.028 & 10 & 0.02789 & 0.39\% &  \\
            & 0.028 & 11 & 0.02771 & 1.03\% &         \\
        2.6 & 0.028 & 12 & 0.02775 & 0.91\% &   0.82\%       \\
            & 0.035 & 13 & 0.03456 & 1.25\% &         \\
            & 0.035 & 14 & 0.03472 & 0.81\% &         \\
            & 0.035 & 15 & 0.03481 & 0.53\% &         \\
            \hline
         & 0.335 & 16 & 0.32900 & 1.79\% &   \\
        5.2& 0.335 & 17 & 0.32743 & 2.26\% &  2.03\%       \\
            & 0.335 & 18 & 0.32817 & 2.04\% &         \\
        \bottomrule
    \end{tabular}
    \caption{Measurement results of different shapes of piles with different measuring area size and different pile volumes.}
    \label{tab:area_grain_types}
\end{table*}


In the setup of the site and equipment, a long-strip experimental site as shown in Fig.\ref{fig:RealPileAndCamera}for corn grain piles is constructed. 
The camera device used for point cloud acquisition is a Microsoft Azure Kinect ToF camera. 
A fixed horizontal sliding rail system is installed above the corn pile, accommodating both handheld device operation for point cloud collection and automated fixed-speed acquisition. 
In the following evaluations, the point cloud measurement dataset includes three different ground area size and 18 different pile shapes as shown in Tab.\ref{tab:area_grain_types}. 
Each specification features corn piles of varying volumes, with three different pile shapes set for each volume.


\begin{figure*}[t]
    \centering
    \includegraphics[width=0.7\linewidth]{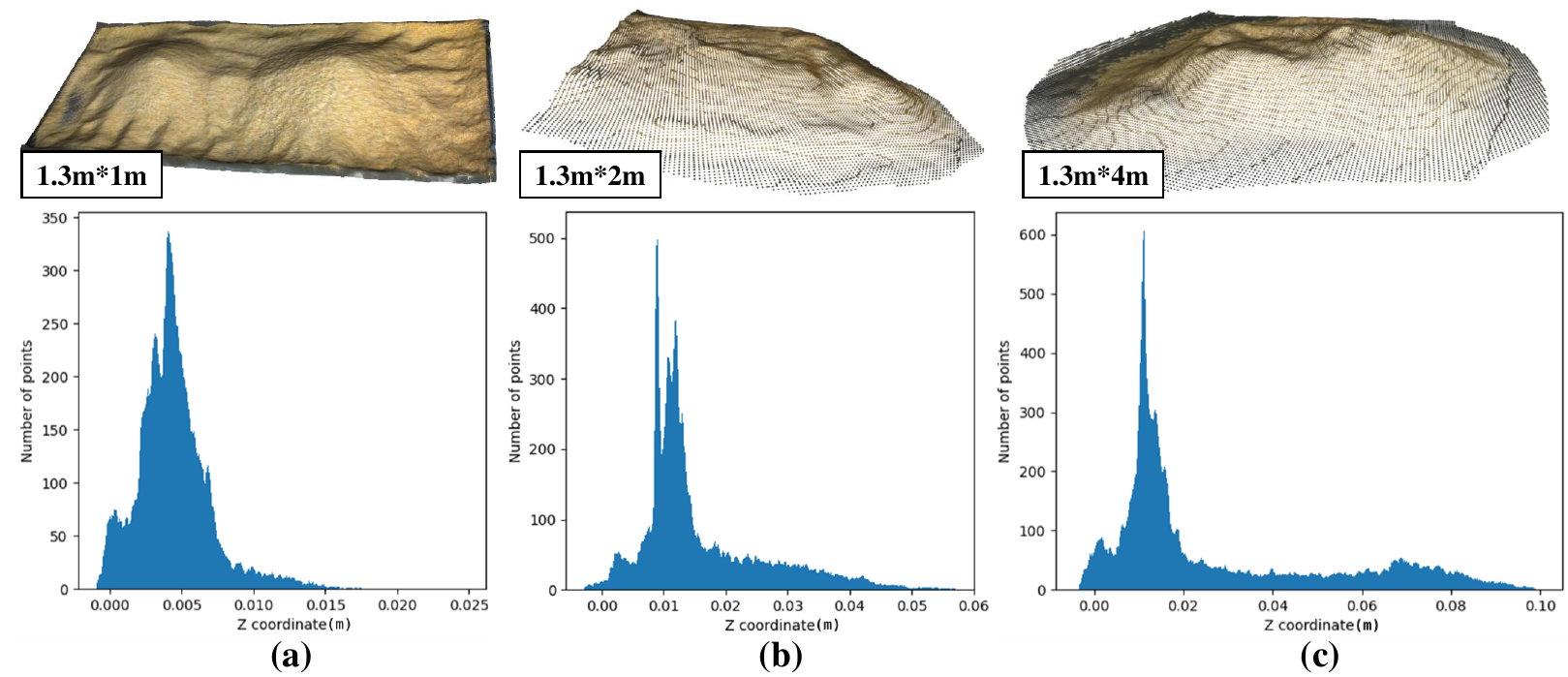} 
    \caption{Different volumetric pile types and height density distribution.}
    \label{fig:PileAndHeight}
\end{figure*}

\begin{figure*}[t]
    \centering
    \includegraphics[width=0.8\linewidth]{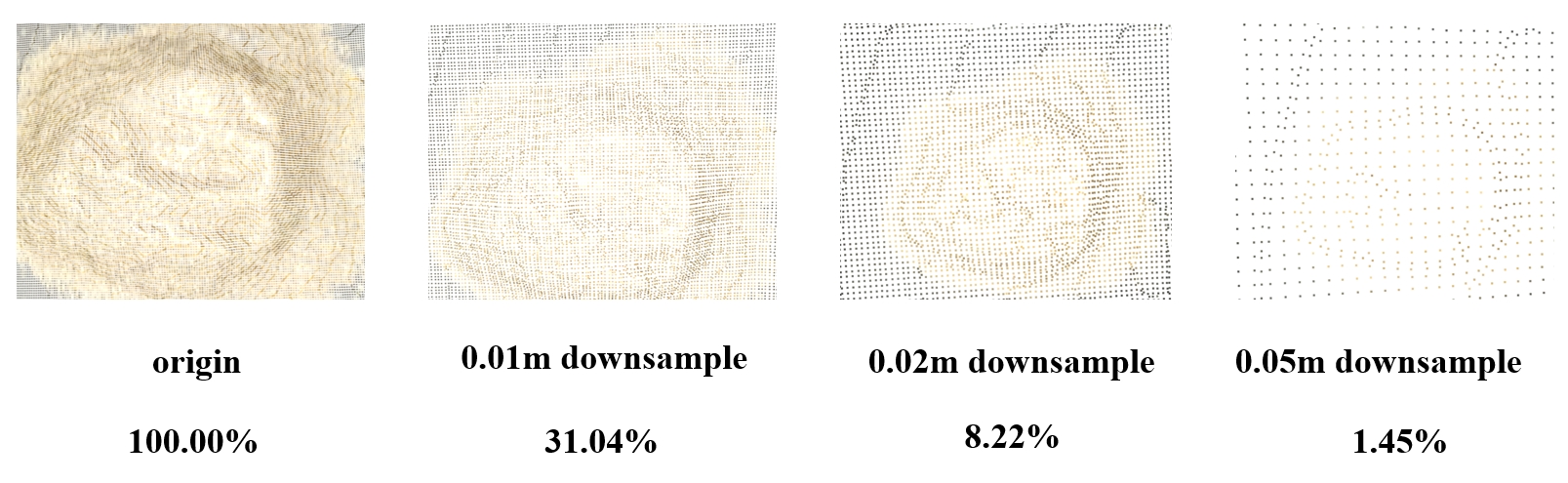} 
    \caption{Pile with different down-sample compressed ratio.}
    \label{fig:downsample}
\end{figure*}

\begin{table*}[h]
    \centering
    \begin{tabular}{lcccccccc}
        \toprule
        voxel size/m & origin & 0.01 & 0.02 & 0.05 & 0.1 & 0.2 & 0.25 & 0.3 \\
        \midrule
        Compressed ratio& 1 & 0.3104 & 0.0822 & 0.0145 & 0.0038 & 0.0011 & 0.0007 & 0.0005 \\
        mean error& 2.03\% & 1.68\% & 2.55\% & 3.17\% & 5.72\% & 8.54\% & 26.18\% & 29.45\% \\
        \bottomrule
    \end{tabular}
    \caption{The algorithm maintains good accuracy after compression to less than 10\%.}
    \label{tab:compression_error}
\end{table*}

The results of automated volume measurements using DIVESPOT on grain piles of varying volumes are shown in Tab.\ref{tab:area_grain_types}. For each pile configuration, 50 round estimations are taken, and the relative errors compared to the ground truth are calculated. It can be observed that in scenarios with a ground area ranging from 1.3 $m^2$ to 5.2 $m^2$, our method maintained a relative error of 2\% or less, demonstrating its robust effectiveness.

Additionally, we conduct down-sampling compression tests on the pile configurations to facilitate application in larger scenarios. The results in Fig.\ref{fig:downsample} and Tab.\ref{tab:compression_error}show that even with compression down to below 1\%, the measurements still achieve approximately 5\% relative error, and with compression down to 0.1\%, the relative error remained within 10\%. This remarkable performance is attributed to the outstanding effectiveness of the ground precise calibration algorithm, which will be further discussed in the subsequent evaluation and discussion sections.

\subsection{Ablation Experiments}

By controlling variables in the volume estimation process, including point cloud filtering, pose correction, and ground calibration, ablation experiments are conducted to validate the effectiveness and role of each component. The contribution of each part to the final volume calculation, as well as their interdependencies, will be verified and discussed in this section based on experimental data.
\begin{table}[h]
    \centering
    \small
    \begin{tabular}{lcc}
        \toprule
        Method& Mean Error& Mean square error\\
        \midrule
        DIVESPOT& 2.03\%& 3.7729e-5 \\
        DIVESPOT-Filter& 5.75\%& 9.7947e-4 \\
        DIVESPOT-Ground calibration& 28.99\%& 8.9829e-3 \\
        \bottomrule
    \end{tabular}
    \caption{Measurements after removal of filtering and ground calibration, the filtering makes the measurements more stable, while the ground calibration accuracy contributes significantly to the measurement accuracy.}
    \label{tab:results}
\end{table}

According to the experimental results in Tab\ref{tab:results}, removing the first radius filtering, the second radius filtering, or both resulted in a change in the method's error from the original 2.03\% to 5.75\%. This indicates that the impact on the method's performance is not substantial; however, the calculation fluctuations under the same volume point cloud significantly increased, with the average variance rising from 3.77e-5 to 9.79e-4. On the other hand, removing pose correction led to a dramatic increase in method error to 28.99\%, with the variance also expanding to 8.40e-3. This simultaneously compromised the method's accuracy and stability, significantly impacting its performance.

These results indicate that the pose correction component is crucial to the method's performance. The original point cloud poses exhibit certain deviations, and the absence of pose correction affects the histogram analysis's ability to accurately determine the ground peak. An inclined pose causes the ground to occupy a larger range in the z-direction, potentially including non-ground parts within this range. Selecting any point within this range as the ground would result in an inaccurately large or small calculation range.From one perspective, for the measured pile-shaped objects, the closer to the ground, the larger the volume represented by each unit height, making the volume measurement more sensitive to ground changes. From another perspective, in practical engineering, to improve the efficiency of traversing the point cloud at each step without applying positive value judgment, point cloud coordinates below the ground are also included in the volume calculation. This inclusion of negative values in the integration results in additional volume reduction, thereby amplifying the sensitivity of volume calculation to the ground position.

\subsection{Comparison of  Volume Calculation Methods}

In our volume estimation scheme, we test three different volume calculation methods using the same filtered and pose-corrected point cloud data set. Given the significant impact of different hyperparameters on the volume measurement methods based on convex hull and slice, we also select different hyperparameter gradients for testing.

\begin{figure}[h]
    \centering
    \includegraphics[width=\linewidth]{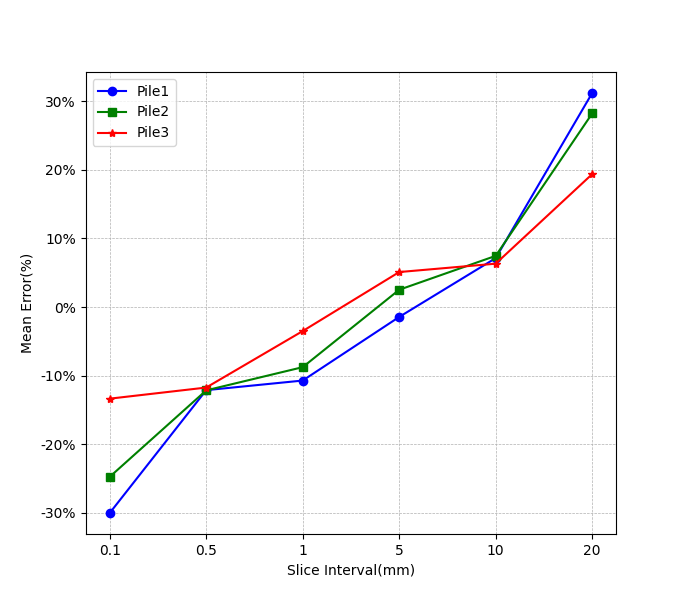} 
    \caption{Difficulty in controlling the accuracy of Slice-based methods with different pile shapes and interval values.}
    \label{fig:slice_experiment}
\end{figure}

As shown in Fig.\ref{fig:slice_experiment}, for the three types of piles with the same volume, the volume estimated using the slice-based method varies significantly with different slice intervals. When the interval is small, the method often does not sufficiently fit within the layers, resulting in an underestimated volume. Conversely, with a larger interval, the method overestimates the volume due to excessive height integration. Within the interval range close to the true value, it is challenging to find a feasible value that stabilizes the calculation across the three types of piles.

\begin{figure}[h]
    \centering
    \includegraphics[width=0.8\linewidth]{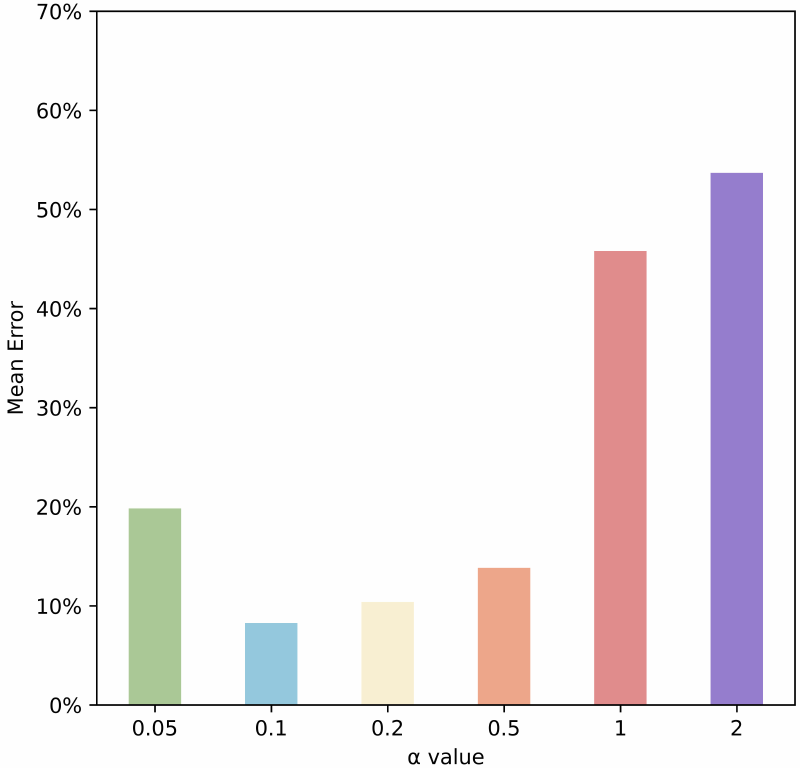} 
    \caption{Accuracy of convex packet-based methods depends on the $\alpha$ value.}
    \label{fig:alpha2meanerror}
\end{figure}

Similarly, as shown in Fig.\ref{fig:alpha2meanerror}, the accuracy of the volume estimation using the convex hull method varies with different $\alpha$ values. This variation is primarily due to differences in surface reconstruction accuracy, with larger $\alpha$ values leading to significant loss of point cloud surface information. These two volume calculation methods heavily depend on parameters and cannot reduce the relative measurement error to 2\% or lower, as effectively as the voxel-based method through ground calibration.

These observations highlight the challenges in parameter sensitivity for both slice-based and convex hull-based methods, emphasizing the need for careful parameter tuning and the advantages of the voxel-based approach in achieving lower estimation errors through ground calibration.

\section{DISCUSSION}

\subsection{Computational Efficiency Optimisation and Testing}

Measurement of piles typically needs to be deployed on edge computing devices. In common industrial measurement scenarios, point cloud reconstruction files usually occupy several gigabytes of storage space, imposing requirements on the algorithm's spatial complexity. Additionally, when the algorithm is deployed on edge devices for estimating calculations, the computational time efficiency, while not requiring high real-time performance, should still complete the corresponding calculation tasks within a relatively short time frame.

\begin{figure}[h]
    \centering
    \includegraphics[width=0.8\linewidth]{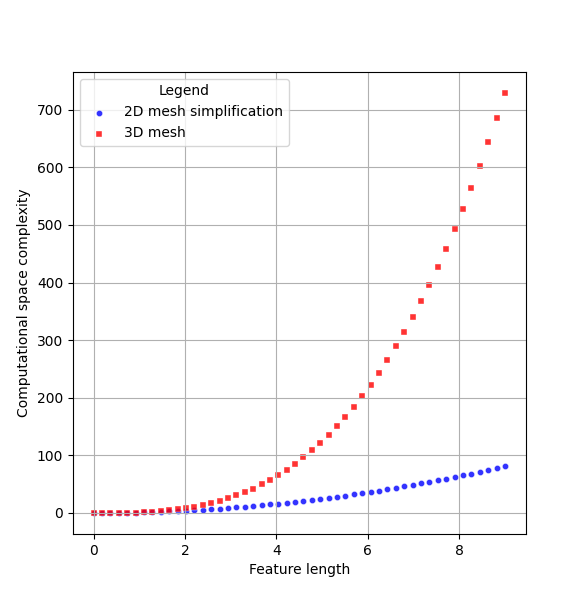} 
    \caption{The 2D projection approximation can optimise the algorithm's demand on device storage as the scene grows larger.}
    \label{fig:space_optimize}
\end{figure}

In terms of storage space optimization, we improve the voxel network-based measurement scheme by approximating the original method of calculating internal space grid points. By leveraging the characteristic that the vertical component of the surface normal vector of pile objects is positive, we transforme the method to multiply the ground grid size by the voxel height and then integrate. As shown in Fig.\ref{fig:space_optimize}, under the same pile slope angle, when the scene volume increases, the method of directly calculating the number of internal voxel network points is related to the total volume of the pile. The total volume is proportional to the base area and pile height. Consequently, the computational spatial complexity is proportional to the cube of the pile's feature length. In contrast, our method simplifies this by integrating the ground grid, resulting in computational spatial complexity proportional to the ground area, which is proportional to the square of the feature length. This approach effectively mitigates the exponential increase in computational space as the scene scale enlarges.

\begin{figure}[h]
    \centering
    \includegraphics[width=0.8\linewidth]{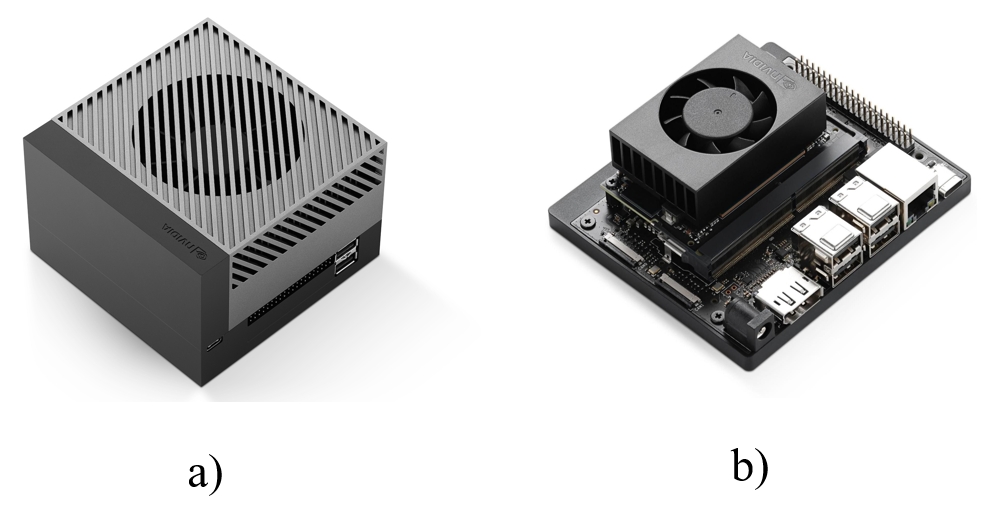} 
    \caption{a) Jetson orin with higher arithmetic power and b) Jetson nano with lower arithmetic power.}
    \label{fig:jetson}
\end{figure}

\begin{table}[h]
    \centering
    \begin{tabular}{lcc}
        \toprule
        Board & Jetson Orin & Jetson Nano \\
        \midrule
        Running time & 86.3s& 432.9s\\
        \bottomrule
    \end{tabular}
    \caption{Volumetric computation of high-precision reconstruction results of 300,000 point clouds on two edge computing devices, with time-consumption that can meet the needs of practical applications}
    \label{tab:running_time}
\end{table}

To test the computational efficiency of the scheme on embedded devices and verify its feasibility for large-scale deployment on edge devices, we deploy the algorithm on the NVIDIA Jetson embedded devices as Fig.\ref{fig:jetson}. The test point cloud cover an area of 5.2 square meters with an initial point cloud size of over 300,000 points. As demonstrated 
 in Tab.\ref{tab:running_time},on the more powerful Jetson Orin, the complete volume calculation took 86.3 seconds, while on the less powerful Jetson Nano, the entire process took 432.9 seconds. Considering a reasonable compression of the point cloud to 5\%, a single edge computing device could handle volume measurement tasks for an area of 100 square meters. By reasonably partitioning the point cloud and employing distributed computing, the requirements for large-scale deployment can be met.

\subsection{Direct Calibration Using External Information}

In the Evaluation section, we assess the importance of ground calibration. Directly using the camera's height above the ground in the real world to calibrate the ground position in the point cloud is also a feasible approach. A fixed rail at a height of 1450 mm is set up to capture the point cloud below it. Based on the fixed point height of 36 mm from the camera lens, a reference height of 1414 mm is obtained as the lens reference height. This height is used as the baseline to calibrate the ground in the point cloud coordinate system. The results of this calibration are shown in Tab.\ref{tab:realheight}.

\begin{table}[h]
    \centering
    \begin{tabular}{lccc}
        \toprule
        Area size/$m^2$ & 1.3 & 2.6 & 5.2 \\
        \midrule
        original method & 1.97\% & 0.82\% & 2.03\% \\
        Calibrated with real height & 18.16\% & 14.78\% & 13.97\% \\
        \bottomrule
    \end{tabular}
    \caption{Guiding the ground calibration by the real height results in large errors}
    \label{tab:realheight}
\end{table}
It can be observed that directly using this external real-world measurement data to form a closed-loop calibration results in systematic errors. Even though our measurement of the lens height exceeds the resolution of the test ToF camera, the results still exhibit these errors. The reason for this unsatisfactory result is that the depth information from the camera itself contains inherent errors. These errors are present in each frame and tend to neutralize when depth images are registered into a point cloud. However, while the point cloud maintains relative positional accuracy, its absolute position is unlikely to align with the original real-world values. Consequently, even when using high-precision external data to calibrate the ground, satisfactory results are difficult to achieve.

\subsection{When The Ground Becomes Uneven}

The premise of simplifying computations through ground grid elements assumes that the ground is flat. However, measures must be taken to account for situations where the ground is uneven. This unevenness can be due to two factors: the actual uneven ground in the real-world scenario and systematic errors introduced during point cloud registration.

When the ground in the actual scenario is uneven, such as sloped or stepped sections, pre-calibration can be performed to provide approximate equations for the corresponding positions, thereby correcting issues caused by the terrain.

Another type of ground unevenness occurs when the ground, originally flat in the real world, becomes uneven in the reconstructed point cloud space. As the point cloud scene scale increases, registration errors can cause originally flat ground in the real-world coordinate system to appear uneven across different regions. In the world coordinate system of the point cloud space, this results in an uneven height distribution of the originally flat ground. In the height density distribution graph, the low-position peak values may become less distinct or even turn into a series of plateau values. To address this, semantic-based registration correction or better registration algorithms should be used\citep{Maleki-23}\citep{10.1115/1.4053272}\citep{10160412}.

A more practical engineering approach is to still extract the lower part of the height density graph. Given that the registration errors are likely random, these deviations can be assumed to follow a Gaussian distribution. Simplifying this, the midpoint of the series of ground peak values can be used as the ground reference value. The results of this method are shown in Tab.\ref{tab:midpeak}.

\begin{table}[h]
    \centering
    \begin{tabular}{lccc}
        \toprule
        Area size/$m^2$& 1.3 & 2.6 & 5.2 \\
        \midrule
        original method & 1.97\% & 0.82\% & 2.03\% \\
        using peak midpoint & 2.99\% & 2.39\% & 4.94\% \\
        \bottomrule
    \end{tabular}
    \caption{Approximation by intermediate peaks also gives a relatively feasible measurement accuracy}
    \label{tab:midpeak}
\end{table}
After such approximate processing, although the average relative error decreases across the three different pile types with varying ground areas used in the experiments, it remains within 5\%. From a practical application perspective, this still constitutes a feasible and straightforward technical remedial measure.

\section{RELATED WORKS}
\textbf{The methods for acquiring 3D scene reconstruction} primarily include those based on RGB images with pose information and neural networks\citep{anciukevivcius2023renderdiffusion}\citep{9857384}\citep{10.1145/3477314.3506998}, as well as those directly based on depth cameras\citep{Hu-20}\citep{rs15184604}\citep{zhou2021research}\citep{Yan-22}. The former currently has two main approaches: Neural Radiance Fields (NeRF)\citep{mildenhall2020nerfrepresentingscenesneural} is a neural network-based technique used to generate high-quality 3D representations from a limited set of 2D images. It works by learning a neural network that maps the input camera poses and image pixels to a 3D space's color and density fields. 3D Gaussian Splatting (3DGS) \citep{kerbl20233dgaussiansplattingrealtime}is a method that represents each point in a point cloud as a three-dimensional Gaussian distribution. The central position, shape, and scale of each point are modeled through a Gaussian distribution, providing a continuous spatial representation. This method allows for smooth interpolation and transition during rendering, generating high-quality images while offering computational efficiency advantages when processing and storing large point cloud datasets. The method of direct acquisition through depth cameras involves capturing single-frame depth information with a depth camera to derive point clouds. This approach matches multiple image information, estimates camera poses, and processes multi-view images to obtain the target point cloud. The general workflow includes automatic pose estimation of the input images and multi-frame matching techniques, ultimately outputting a densified point cloud. In comparison, both methods rely on captured pose information. However, the neural network-based methods depend more on sample training and RGB color information, whereas depth cameras have lower requirements for shooting conditions.

\textbf{Point cloud volume calculation methods} can be categorized into three principles: slice-based methods\citep{stefanidou2020lidar}\citep{rs15205006}\citep{doi:10.1080/01431161.2018.1541111}\citep{isprs-archives-XL-5-101-2014}\citep{10194919}, voxel-based methods, and convex hull-based methods. The slice-based method involves dividing the object's point cloud into layered slices and calculating the total volume by integrating the volumes of the local slice point clouds. The voxel-based method extends the concept of calculating two-dimensional areas to three-dimensional volumes\citep{kang2020review}. This method divides the three-dimensional space into a regular 3D voxel grid and calculates the volume by counting the globally consistent voxels\citep{wang2020estimation}\citep{ZHAO2021106131}. The convex hull-based method constructs a convex hull structure around the point cloud and calculates the volume inside the convex hull to obtain the point cloud's volume.The convex hull method has two branches: three-dimensional convex hull and two-dimensional convex hull. The three-dimensional convex hull\citep{griwodz2021alicevision} \citep{https://doi.org/10.1155/2024/6662678}generally uses the $\alpha$-shape method (also known as the rolling ball method) to construct a surface triangle network, establish a closed convex hull of the object, then calculate the positive and negative volumes of each triangular prism projected onto a given plane, and sum them to obtain the object's volume. The two-dimensional convex hull is mainly used in conjunction with the slice-based method, where the shape of each slice is optimized using the convex hull method after slicing, and then the area of the slice is calculated. The voxel-based method measures volume by simply counting the number of voxels within the measured object's spatial grid.The application of point cloud volume estimation for irregular objects spans various fields such as transportation, animal husbandry, forestry, and mining\citep{leite2020estimating}\citep{rs15205006}\citep{10160658}\citep{agriculture13071405}. Laser scanning technology can obtain detailed point cloud data of road surfaces, and algorithms can be used to reconstruct three-dimensional models of potholes to calculate their volumes\citep{10.1117/12.2573129}. For animals such as livestock, an improved Poisson reconstruction algorithm can process point cloud data of pigs to reconstruct surface features and calculate their volume, which can further be used to estimate livestock volume\citep{ani14081210}. This method overcomes the issues of irregular point cloud shapes and partial data loss, ensuring the continuity and completeness of the reconstruction results.For irregular fruits, a consumer-grade RGB-depth sensor can be used for the three-dimensional reconstruction and volume estimation of dates. By collecting point cloud data and combining it with depth images, a three-dimensional model of the dates is generated to calculate their volume and surface area\citep{10149328}. Model-based fitting and reconstruction can also be applied, as demonstrated in the measurement of potato volume using laser triangulation and 3D reconstruction technology. Laser scanning obtains the point cloud data of potatoes, and interpolation and reconstruction algorithms generate their 3D model to calculate volume and predict weight\citep{9206528}.

\section{CONCLUSION}

For addressing the common problem of volume measurement for pile-type objects in actual industrial scenarios, we propose DIVESPOT, a 3D pile volume estimation scheme based on point clouds. In experiments involving 18 different pile shapes across three scenarios with varying ground areas of corn heaps, DIVESPOT achieves a relative estimation error of less than 2\%. DIVESPOT automates point cloud filtering and pose correction, and it rapidly and accurately identifies and segments the ground without relying on sample-based training or RGB color information. It has proven to be highly adaptable to compressed point clouds and deployment on edge computing devices. This research provides an effective approach for the volume estimation of pile-type objects in industrial applications and offers a comprehensive measurement system solution.

\bibliographystyle{ACM-Reference-Format}
\bibliography{reference}
\end{document}